\documentclass[lettersize,journal]{IEEEtran}
\usepackage{amsmath,amsfonts}
\usepackage{algorithmic}
\usepackage{algorithm}
\usepackage{array}
\usepackage[caption=false,font=normalsize,labelfont=sf,textfont=sf]{subfig}
\usepackage{textcomp}
\usepackage{stfloats}
\usepackage{url}
\usepackage{xcolor}
\usepackage{verbatim}
\usepackage{multirow}
\usepackage{makecell}
\usepackage{graphicx}
\usepackage{cite}
\begin{document}

\title{EEsizer: LLM-Based AI Agent for Sizing of Analog and Mixed Signal Circuit}

\author{
	\vskip 1em
	
	Chang Liu, and Danial Chitnis 

	\thanks{
		Manuscript received XX, XXXX; revised XX XXXX, XX; accepted XXXX XX, XXXX.
		Chang Liu is sponsored by Peter Denyer's PhD Scholarship at The University of Edinburgh
		
		The authors are with The School of Engineering, Institute for Integrated Micro and Nano Systems, University of Edinburgh, Edinburgh, EH9 3FF, UK.
		
        (corresponding author: Danial Chitnis;  e-mail: d.chitnis@ed.ac.uk).
	}
}



\maketitle

\begin{abstract}
The design of Analog and Mixed-Signal (AMS) integrated circuits (ICs) often involves significant manual effort, especially during the transistor sizing process. While Machine Learning techniques in Electronic Design Automation (EDA) have shown promise in reducing complexity and minimizing human intervention, they still face challenges such as numerous iterations and a lack of knowledge about AMS circuit design. Recently, Large Language Models (LLMs) have demonstrated significant potential across various fields, showing a certain level of knowledge in circuit design and indicating their potential to automate the transistor sizing process. 

In this work, we propose EEsizer, an LLM-based AI agent that integrates large language models with circuit simulators and custom data analysis functions, enabling fully automated, closed-loop transistor sizing without relying on external knowledge. By employing prompt engineering and Chain-of-Thought reasoning, the agent iteratively explores design directions, evaluates performance, and refines solutions with minimal human intervention.

We first benchmarked 8 LLMs on six basic circuits and selected three high-performing models to optimize a 20-transistor CMOS operational amplifier, targeting multiple performance metrics, including rail-to-rail operation from 180 nm to 90 nm technology nodes. Notably, OpenAI o3 successfully achieved the user-intended target at 90 nm across three different test groups, with a maximum of 20 iterations, demonstrating adaptability and robustness at advanced nodes. To assess design robustness, we manually designed a bias circuit and performed a variation analysis using Gaussian-distributed variations on transistor dimensions and threshold voltages. Overall, the results demonstrate the potential of LLMs to accelerate AMS circuit design by reducing manual effort while maintaining efficiency and adaptability across circuits, models, and technology nodes.

\end{abstract}


\section{Introduction}
\begin{figure}[t]
    \centering
    \includegraphics[width=0.5\textwidth]{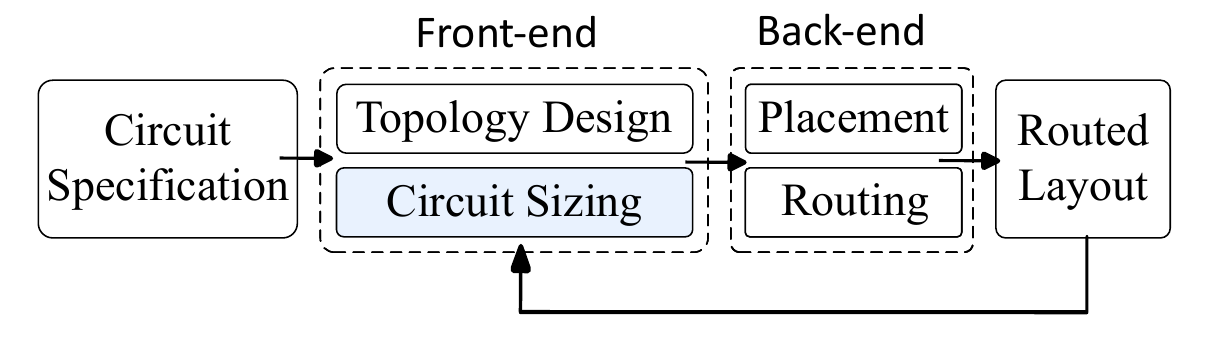}
    \caption{A typical circuit design flow. The process begins with front-end stages including topology selection and circuit sizing, followed by back-end tasks such as placement, routing, and post-layout simulation. The results from post-layout simulation often necessitate adjustments to the circuit sizing.}
    \label{fig:design_flow}
\end{figure}

\IEEEPARstart{A}{nalog} and mixed-signal (AMS) integrated circuits, such as Analog-to-Digital Converters (ADCs) \cite{761034}, filters \cite{9245137}, and Power Management Integrated Circuits (PMICs) \cite{ballo2021review}, play a vital role in communication systems and consumer electronics. The design process of AMS circuits typically follows a structured flow, beginning with the front-end stages of topology selection and circuit sizing, followed by the back-end tasks of placement and routing, as illustrated in Fig.~\ref{fig:design_flow}. Throughout this process, designers must often consider trade-offs among Power, Performance, and Area (PPA), to achieve a balanced circuit performance. Among these stages, circuit sizing remains particularly challenging due to its high-dimensional design space and complex trade-offs between performance metrics. Even for a fixed circuit topology, different applications impose distinct target specifications, and each metric has its own sensitivity and sizing profile. As a result, sizing is an iterative process rather than a one-time task, balancing competing objectives across varying requirements \cite {10.1145/3569052.3578929}. Therefore, as a crucial aspect of Electronic Design Automation (EDA), automatic circuit sizing has garnered increasing research interest \cite{lyu2018batch}, \cite{liao2017parasitic}.

The rapid expansion of Machine Learning (ML) techniques has increasingly highlighted its potential for automating circuit design, offering new possibilities for achieving automation in the circuit sizing process. Bayesian Optimization (BO) frames transistor sizing as an optimization problem in which the objective function is evaluated through simulation, and uses Gaussian Process (GP) regression to efficiently guide the search toward high-performing solutions \cite{lyu2018batch, touloupas2021local}. Deep Reinforcement Learning (DRL) further improves sizing efficiency by leveraging rewards from iterative simulations, enabling adaptation to complex design spaces \cite{zhao2022analog}. However, these methods operate without domain-specific analog design knowledge, leading to high computational costs and inconsistent performance across varying circuit designs and performance metrics. \cite{settaluri2020autockt}.

Recently, large language models (LLMs) such as GPT-4 \cite{achiam2023gpt}, Llama~3 \cite{touvron2023llama} and the OpenAI~o-series \cite{jaech2024openai} have demonstrated exceptional capabilities in tasks including natural language processing, code generation, and reasoning. All of these LLMs utilize transformer architectures with self-attention mechanisms and are pre-trained on large, diverse text corpora. This pre-training enables them with knowledge relevant to circuit design \cite{yin2024ado}, in-context learning capabilities \cite{dong2022survey}, and decision-making skills \cite{li2022pre}. Leveraging these abilities, LLMs can propose circuit optimization strategies and select appropriate functions to analyze simulation data, enabling automated AMS circuit design. Nevertheless, several challenges remain. First, LLMs have a limited understanding of AMS circuit topologies, which makes it challenging to capture trade-offs that affect circuit performance. Second, the lack of high-quality open-source analog design data limits opportunities for training and benchmarking. Finally, their constrained mathematical precision reduces reliability in accurate sizing and performance estimation, necessitating seamless integration with external tools such as circuit simulators and data analysis functions.

\begin{table}[t]
\centering
\caption{Comparison of recent LLM-based approaches for analog circuit design, highlighting their sizing optimizers, supported performance metrics, node transfer ability, and topology generation capabilities.}
\renewcommand{\arraystretch}{1.3}
\resizebox{0.5\textwidth}{!}{ 
\begin{tabular}{c|c|c|c|c}
    \hline
    \textbf{Method} & \makecell{\textbf{Sizing} \\ \textbf{Optimizer}} & 
    \makecell{\textbf{Perf.} \\ \textbf{Metrics}} &  
    \makecell{\textbf{Node} \\ \textbf{Transfer}} &  
    \makecell{\textbf{Topology} \\ \textbf{Gen.}} \\
    \hline
    \hline
    AmpAgent\cite{liu2024ampagent}   & \makecell{ABC\cite{article} \\ TuRBO\cite{eriksson2019scalable}} & 4 & No & Yes \\
    Artisan\cite{chen2024artisan}    & Gm/Id scripts\cite{8742812} & 5 & No & Yes \\
    Atelier\cite{shen2025atelier}    & CMA-ES\cite{Hansen} & 7 & No & Yes \\
    LEDRO\cite{kochar2025ledro}      & TuRBO & 4 & Yes & No \\
    LLMACD\cite{11101244}            & TBM\cite{zhi2025analog} & 4 & No & No \\
    LLM-USO\cite{somayaji2025llm} & BO & 4 & No & No \\
    EEsizer \cite{11107079} & LLM & 9 & No & No\\
    \hline
\end{tabular}
}
\label{tab:related works}
\end{table}

To address the problems, several studies have explored the use of LLMs for automated analog circuit design, focusing on circuit design knowledge and sizing optimizer integration to support topology generation and transistor sizing. Artisan enhances the Llama2-7b model with specialized opamp design knowledge, employing a bidirectional circuit representation to bridge netlists and natural language, and utilizes gm/Id methodology for transistor sizing \cite{8742812}. AmpAgent employed retrieval-augmented generation (RAG) to address LLM's knowledge gaps in amplifiers and use conventional optimization algorithms to size the transistors \cite{liu2024ampagent}. Atelier integrates LLMs with a knowledge base, while utilizes Covariance Matrix Adaptation Evolution Strategy(CMA-ES) to optimize a broader range of performance metrics, including noise and area \cite{chen2024artisan}. LEDRO leverages LLMs to iteratively shrink the design space, improving Trust Region Bayesian Optimization (TuRBO)’s efficiency and enabling technology nodes transfer \cite{kochar2025ledro}. Other approaches include LLMACD, which applies a Transistor Behavioral Model (TBM) to directly optimize behavioral parameters that impact performance \cite{11101244}, and LLM-USO, which encodes circuit design knowledge in structured text and uses Bayesian Optimization (BO) for sizing \cite{somayaji2025llm}. Our previous work \cite{11107079} employs Chain-of-Thought (CoT) prompting to guide LLMs in sizing a 20-transistor opamp across nine performance metrics. Overall, these approaches illustrate the possibility of LLMs in analog and mixed-signal circuit design, showing promising potential to reduce manual effort, improve efficiency, and expand the scope of automated optimization. However, as summarized in Table~\ref{tab:related works}, existing methods often depend on manually annotated datasets for knowledge retrieval and domain-specific fine-tuning or on external black-box optimizers for transistor sizing, which limits their adaptability to new technology nodes and to a broader set of performance metrics.

In this paper, we extend our previous work \cite{11107079} by updating EEsizer, our LLM-based AI agent for AMS circuit design, focusing on the transistor sizing process at smaller nodes, including detailed evaluations. To address the limitations of previous methods, EEsizer incorporates several design choices. First, we reduce the dependence on large pre-collected datasets by adopting a closed-loop methodology where the LLM actively queries the simulator, generating its own labelled data during the optimization process, which can be adapted across different circuits, specifications and technology nodes. Second, instead of relying solely on the LLM's implicit knowledge of analog circuits, we leverage Chain-of-Thought (CoT) reasoning within a relevant context. This enables the model to capture performance trade-offs, explore design directions, and make informed decisions iteratively within the optimization loop. Third, to compensate for LLMs’ limited mathematical precision, we integrate them with the Ngspice simulator and custom analysis tools by function calling, which enables accurate performance evaluation. 

To evaluate the proposed agent, we first benchmarked eight LLMs on six basic circuits, demonstrating the EEsizer's compatibility across different models, and selected three high-performing models—Gemini 2.0 Flash, GPT-4.1, and OpenAI o3—for further study. We then applied the EEsizer to a 20-transistor CMOS operational amplifier with a complementary input stage and a class-AB output stage. The agent was tasked with achieving ten performance metrics, including rail-to-rail operation, across 180 nm, 130 nm, and 90 nm on Predictive Technology Model (PTM) technology nodes \cite{10.1145/1229175.1229176}. Finally, we manually designed a bias circuit and performed variation tests on both transistor dimensions and threshold voltages to assess the reliability of the optimized circuit.
\section{Methodology}
EEsizer takes a circuit netlist and expected performance metrics as user inputs and ultimately generates an optimized netlist that meets the target performance, along with the reasons for parameter adjustments. 
\begin{figure*}[t]
\centering
\includegraphics[width=1\textwidth]{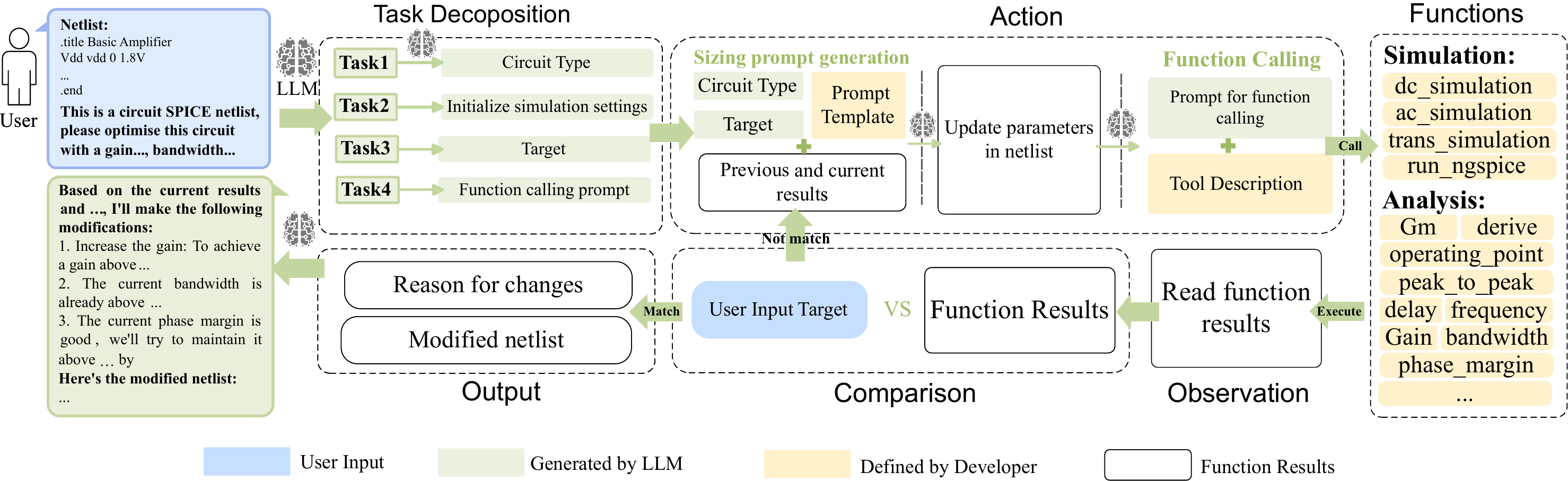}
\caption{The entire process for circuit sizing with EEsizer. The process begins with the task decomposition stage, generating four tasks for different stages. Action, observation and comparison formed a ReAct optimization loop. Finally, the agent generates output consisting of reasons for changes and modifications of the netlist for the user.}
\label{fig:whole process}
\end{figure*}

The entire process for circuit sizing with EEsizer begins with user input. In the optimization process, action, observation, and comparison create a Reasoning and Acting (ReAct) loop \cite{yao2023react}, where the agent cycles through reasoning (analyzing) and taking action based on its insights. It starts with sizing prompt generation. The agent then selects appropriate functions to call external simulator and analyze simulation result based on performance metrics. After executing these functions, it reviews the results and compares them with the user’s input to check for alignment. The final output is produced if the results meet the performance criteria. Ngspice, an open-source SPICE simulator, is integrated into EEsizer to enable in-loop simulation, ensuring transparency and re-producibility through the public release of our source code and benchmark data. The transistor models used in this work are PTM models \cite{10.1145/1229175.1229176}, which are open-source models used for academic and circuit design research. To improve the agent's interpretability, this agent will provide not only the modified netlist but also the reasoning behind each parameter adjustment in every iteration. The entire process for circuit sizing with the EEsizer is shown in Fig.~\ref{fig:whole process}. 

\subsection{Large Language Models}
Recent advancements in large language models (LLMs) have demonstrated strong performance across natural language processing, code generation, and reasoning tasks. These models differ in size, reasoning ability, knowledge, and response speed. Thinking models, such as the o-series, internally generate thinking tokens before answering, often exhibiting higher reasoning ability but longer latency, resulting in slower end-to-end response times \cite{jaech2024openai}. Large-scale models such as GPT-4 and Claude~3.5 provide broad knowledge and generally faster responses compared to thinking models \cite{achiam2023gpt}. Mid-sized models, such as Gemini~2.0~Flash, prioritize faster responses, with a latency of approximately 0.4~s from input to first token, roughly 40 times faster than the OpenAI~o3 model—and 20× lower input/output token expenses. However, Gemini~2.0~Flash offers less reasoning depth than the larger models, which may affect performance on highly complex multi-step tasks \cite{rane2024gemini}.

\subsection{User Input}
The user input serves as the starting point for the optimization process and provides the necessary context for the AI agent to operate effectively. Two key components are required: (1) a circuit SPICE netlist, and (2) a set of target performance metrics specified by the user. The netlist provides a structural description of the circuit, including devices, connections, and initial sizing values, while the performance metrics define the design objectives, including gain, bandwidth, phase margin, delay, or power. 

\begin{figure}[t]
    \centering
    \includegraphics[width=0.5\textwidth]{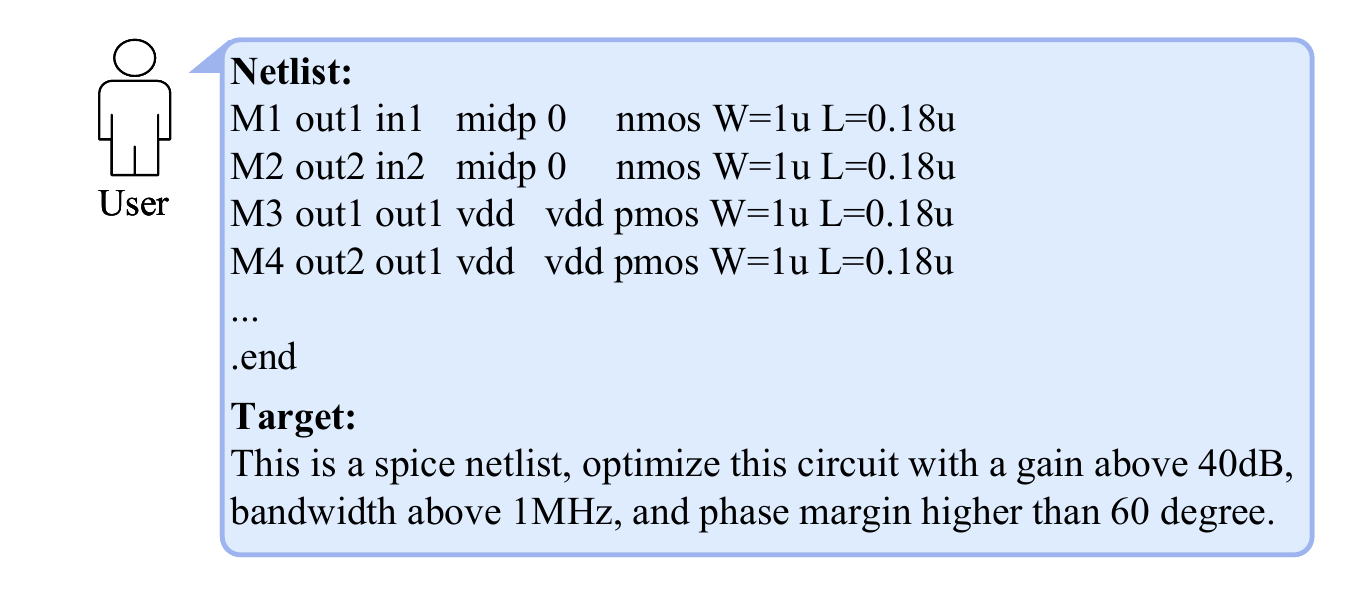}
    \caption{Example of user input consisting of a SPICE netlist that defines the circuit structure, devices, connections, and initial sizing values, and a set of target performance metrics provided by the user, including gain, bandwidth, and phase margin. These two components are the only required inputs for EEsizer.}
    \label{fig:user input}
\end{figure}

Other information, such as circuit type or domain-specific design knowledge, is deliberately excluded from the user input to ensure generality and minimize reliance on manual annotations. Instead, the agent is responsible for inferring circuit characteristics and contextual information during the task decomposition stage. This design choice highlights the automation capability of the EEsizer, allowing it to adapt across different circuit types and technology nodes with minimal human intervention. An example for user input is shown in Fig.~\ref{fig:user input}.

\subsection{Task Decomposition}
Task decomposition divides the circuit sizing process into four sub-tasks, each designed to streamline interaction with the LLM and ensure automated optimization. The task prompts are predefined at the system level to transform the user’s initial query into actionable steps across the flow. Task~1 focuses on circuit identification, where the model recognizes the type and sub-structure of the provided circuit, thereby establishing a relevant context for LLMs. Task~2 extracts input and output nodes, enabling the generation of complete simulation configurations, including sources and measurement points. Task~3 retrieves target performance values from the user input for subsequent comparison. Task~4 maps user-specified requirements, including gain, bandwidth, and power, into structured targets that guide function calls for simulation and data analysis. Together, these sub-tasks provide a modular framework that enhances interpretability, automation, and convergence efficiency in the proposed agent.

\subsection{Chain-of-Thought Prompting}

\renewcommand{\arraystretch}{1.1}
\begin{table}[t]
\centering
\caption{Range of Transistor Size and Bias Voltages for Different Technology Nodes}
\resizebox{0.48\textwidth}{!}{%
    \begin{tabular}{c|c|c|c}
        \hline
         \textbf{\textit{Technology Node}} & \textbf{\textit{180 nm}} & \textbf{\textit{130 nm}} & \textbf{\textit{90 nm}} \\
        \hline
        \hline
        Width   & [0.18, 400] µm  & [0.13, 400] µm   & [0.09, 400] µm \\
        Length  & [0.18, 18] µm   & [0.13, 13] µm    & [0.09, 9] µm \\
        Supply Voltage  & 1.8 V  & 1.8 V   & 1.2 V \\
        Bias Voltage    & (0, 1.8) V  & (0, 1.8) V   & (0, 1.2) V \\
        \hline
    \end{tabular}%
}
\label{tab:range}
\end{table}

\begin{figure}[t]
    \centering
    \includegraphics[width=0.5\textwidth]{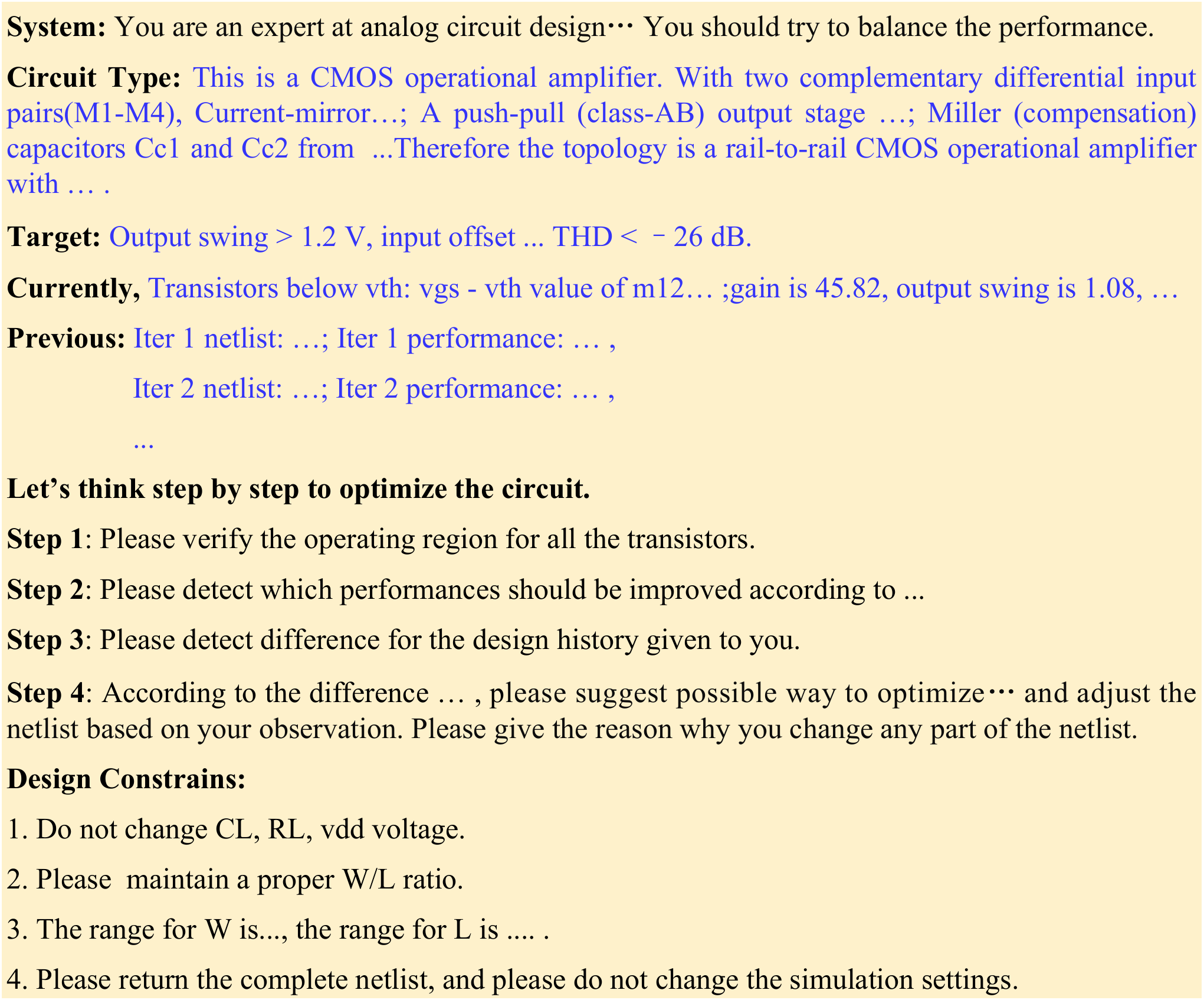}
    \caption{Chain-of-Thought (CoT) guided optimization framework, in which the circuit type, target performance, current results, and historical results, generated by LLMs, are integrated into the context (highlighted in blue), while the underlying template remains in black.}
    \label{fig:context}
\end{figure}

\begin{figure}[t]
    \centering
    \includegraphics[width=0.5\textwidth]{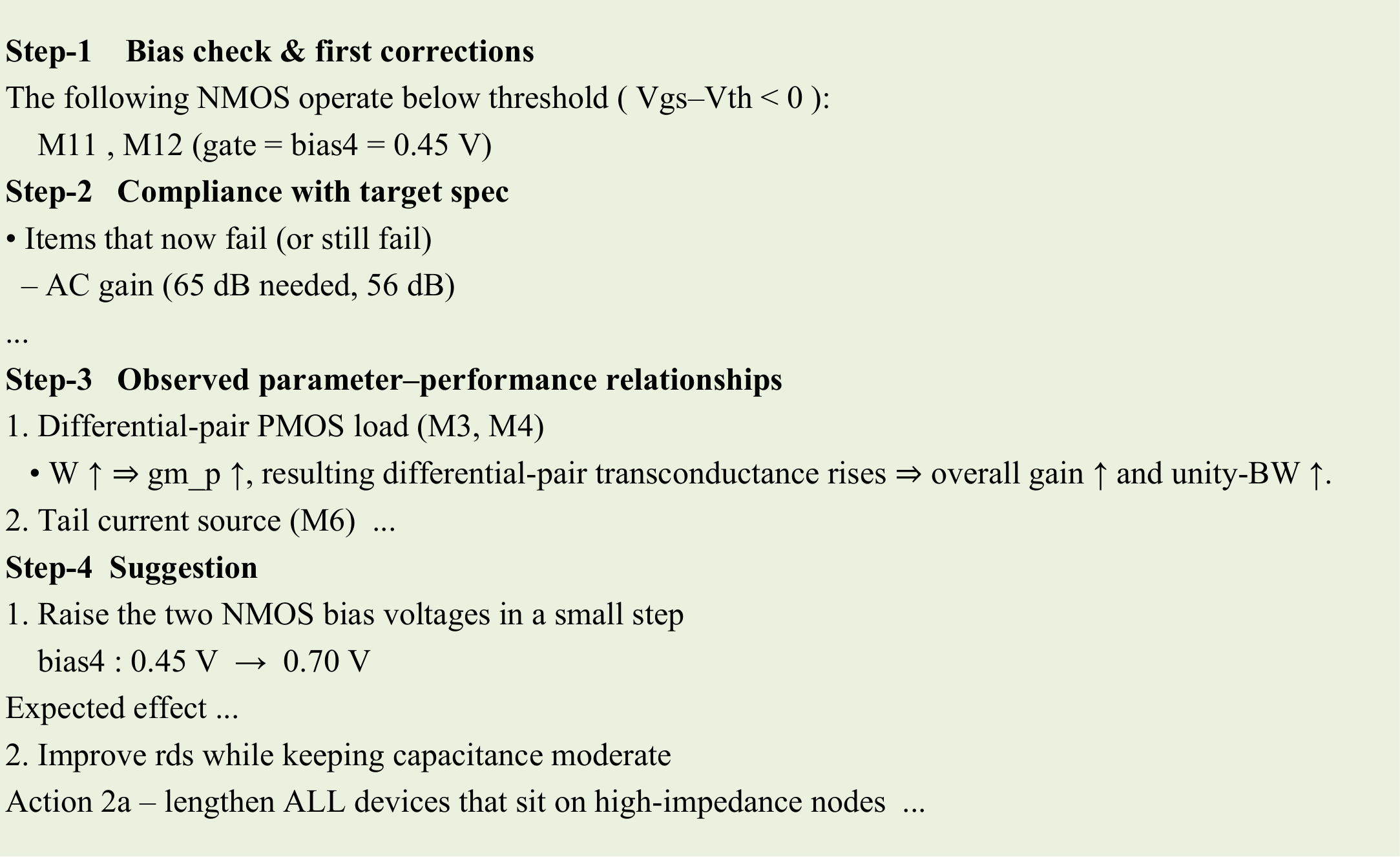}
    \caption{An example of LLM response for the CoT prompt.}
    \label{fig:output}
\end{figure}

In this work, we employ Chain-of-Thought (CoT) prompting \cite{wei2022chain} to guide LLMs in generating structured optimization strategies. While reasoning models, such as OpenAI~o3, are capable of direct problem solving, combining them with CoT allows for step-by-step intermediate reasoning, enabling a more systematic evaluation of constraints and performance metrics, and enhancing both accuracy and efficiency in transistor sizing. The overall template structure is shown in Fig.~\ref{fig:context}.

The prompt template for CoT consists of a system prompt and four main parameters, including the user input target, circuit type, previous results and current results. The system prompt incorporates relevant circuit design knowledge of the LLMs to guide the following process. The circuit type is identified in Task~1. The previous results, current results and target performance provide highly relevant context for LLMs, enabling effective in-context learning, which helps LLMs better understand the relationship between performance and parameters, prompting them to generate new, potentially optimal design points. These points are then simulated, and the results are used to further enrich the context history. The design constraints ensure consistent supply voltage, fixed transistor models, and parameter ranges including width, length, and bias voltage, as summarized in Table~\ref{tab:range}. The CoT instructions include four steps: (1)~verify transistor regions, (2)~identify performances needing improvement, (3)~connect design parameters with performance metrics, and (4)~propose optimization within constraints. The resulting suggestions prompt LLMs to generate new design points, which are simulated and fed back into the context for iterative refinement. An example of LLM response for the CoT prompt is shown in Fig.~\ref{fig:output}.

\subsection{Simulation in-loop}
\begin{figure}[t]
\centerline{\includegraphics[width=0.5\textwidth]{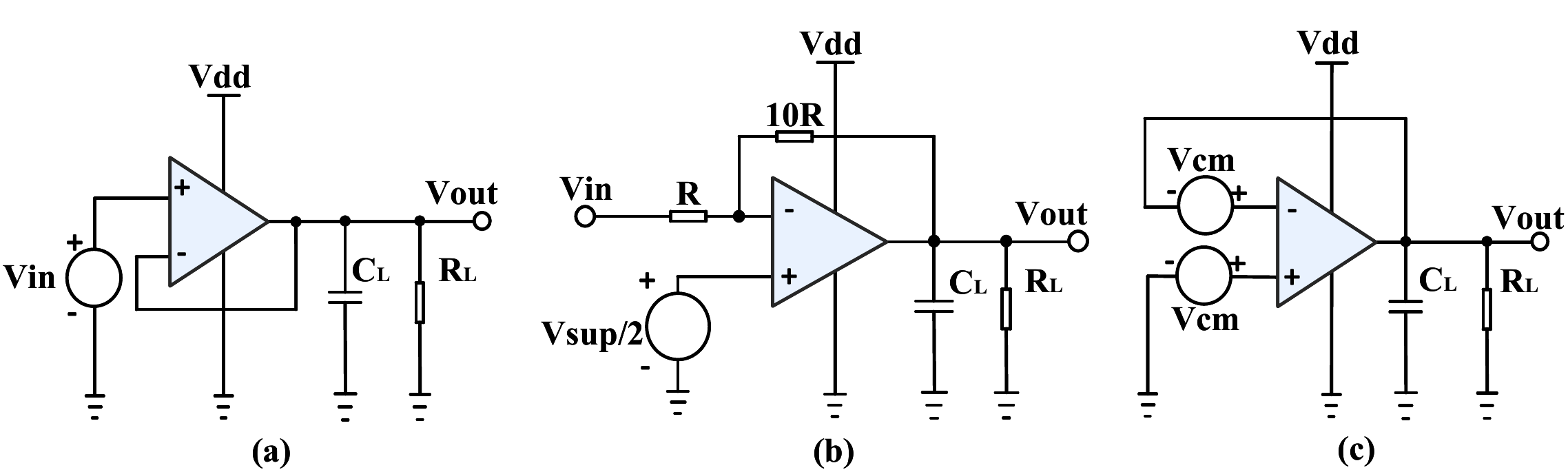}}
\caption{Circuit configurations for measurement: (a)~Unity-gain configuration for offset, ICMR. (b)~Open-loop configuration for output swing, with the non-inverting input fixed at Vsup/2. (c) Configuration for CMRR measurement, sweeping Vcm from 0 to VDD.}
\label{figconfig}
\end{figure}

We set up the simulation in-loop by function calling. Function calling is a structured communication protocol which allows LLMs to interact with external tools, thereby improving their performance in specialized domains. All the functions are pre-defined in the agent by the developer as listed in Table.~\ref{tab:functions}. The configurations for the circuit when measuring a specific performance other than the original are shown in Fig.~\ref{figconfig}.  

In the function calling process, the agent sends a question to the function, which is the Task~4 result, including netlist and target performance, to the LLM. After running the simulation, Ngspice returns the results to the agent, which executes additional functions as needed. The agent finally processes the output and converts it into a natural language response, providing the user with a clear and understandable answer. If the user’s question does not require function calls, the model generates a response as usual. In this case, the LLM is called only once.

After each simulation, the obtained circuit outputs are tested against the required specifications. As shown in the Algorithm. ~\ref{alg:comparison}, this process consists of two main steps: specification comparison and success evaluation.

In the specification comparison stage, the agent evaluates each performance metric, including gain, output swing, input offset, and ICMR, against its corresponding target value within a 5\% tolerance. Each metric is assigned a Boolean flag indicating whether the requirement is satisfied.

\begin{table*}[t]
\renewcommand{\arraystretch}{1.2}
\centering
    \caption{Functions integrated in the agent, organized by circuit type, function type, and configuration when simulating and calculating corresponding performance.}
    \begin{tabular}{|c|c|c|c|c|}
        \hline
        \textbf{\textit{Function}} & \textbf{\textit{Circuit Type}} & \textbf{\textit{Function Type}} & \textbf{\textit{Description}} & \textbf{\textit{Configuration}} \\
        \hline
        \hline
       DC Simulation        & General  & \multirow[c]{3}{*}{Simulation Function}    & \multirow[c]{3}{*}{Add start value, step, end value, output node settings} & \multirow[c]{3}{*}{/} \\ 
        AC Simulation        & General  &                                                                         &                                    &                       \\ 
        Transient Simulation & General  &                                                                         &                                    &                       \\ 
        \hline
        Run Ngspice          & General  & Calling simulator                & Call the external simulator Ngspice to perform simulation & / \\
        \hline
        DC Gain              & Amplifier & \multirow[c]{11}{*}{Analysis Function} & Small signal voltage gain at low frequency (10kHz) & Amplifier \\ 
        Input Offset              & Amplifier   &            & Output error when input = Vsup/2          & unity-gain \\
        ICMR   & Amplifier   &            & The linear part of the close-loop transfer curve           & unity-gain  \\
        THD                       & Amplifier   &             & The total harmonic distortion measured across full swing               & unity-gain \\
        Output Swing              & Amplifier   &            & The linear part of the open-loop transfer curve            & Open-loop \\
        Bandwidth                 & Amplifier   &             & Frequency when gain -3dB                   & Amplifier \\
        Unity-Gain Bandwidth      & Amplifier   &             & Frequency at unity gain         & Amplifier \\
        Phase Margin              & Amplifier   &             & Phase shift from 180° at unity gain               & Amplifier \\
        CMRR                      & Amplifier   &             & The minimum CMRR measured across the whole Vcm range               & Amplifier \\
        
        Power                     & General     &             & Transient power consumption        & Original  \\
       Delay                      & Logic       &             & Time from 50\% input to 50\% output     & Original  \\
        Frequency                 & Oscillator  &             & Oscillation frequency             & Original  \\
       \hline
    \end{tabular}
\label{tab:functions}
\end{table*}

\subsection{Comparison and Iteration}
\begin{algorithm}[H]
\caption{Comparison with User Input Target and Success Evaluation }
\begin{algorithmic}
\STATE {\textsc{CHECK}}$(\mathbf{outputs}, \mathbf{targets})$
\FOR{each specification $s \in \{$gain, output\_swing, input\_offset, icmr, ...$\}$}
    \IF{$s\_target \neq \text{None}$}
        \STATE Compare $s\_output$ with $s\_target$ within tolerance
        \STATE Set $s\_pass \gets \text{True or False}$
    \ENDIF
\ENDFOR
\STATE
\STATE {\textsc{SUCCESS EVALUATION}}$(\mathbf{pass\_flags}, \mathbf{vgscheck})$
\IF{all $s\_pass = \text{True}$ \AND $vgscheck =$ ``No values found where $V_\text{gs} - V_\text{th} < 0 $''}
    \STATE $succeed \gets \text{True}$
\ELSE
    \STATE $succeed \gets \text{False}$
\ENDIF
\STATE \textbf{return} $succeed$

\end{algorithmic}
\label{alg:comparison}
\end{algorithm}
In the convergence check stage, the algorithm determines whether the optimization has successfully met all design targets. Success is declared only if (1)~all specification flags are true and (2)~no transistors are operating in sub-threshold regions ($V_{\text{gs}} - V_{\text{th}} < 0$), which is constrained to strong inversion reduce the effect of variations and the need for extensive Monte Carlo simulations within the optimization loop. If success is achieved, the optimization terminates and provides the final results to the user. Otherwise, the feedback, which includes result history and the latest outputs, is used to guide the next iteration of parameter updates. This creates a dynamic priority system, where the feedback for each iteration is inherently focused on the specifications that have most recently failed, guiding the LLM directly toward the remaining design gaps, thereby converging to a final design solution.

\begin{figure}[t]
\centerline{\includegraphics[width=0.5\textwidth]{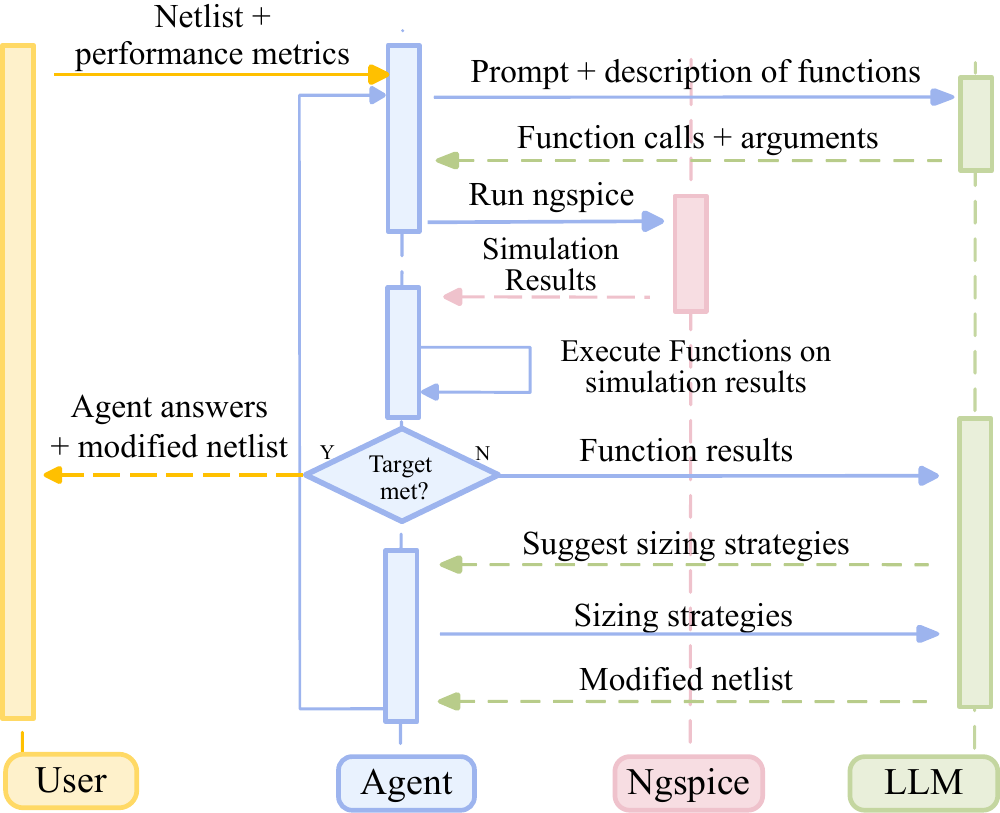}}
\caption{Communication flow between the agent and LLM APIs. The agent constructs structured prompts containing circuit context and optimization history, which are sent to the LLM through its API. The LLM responds with reasoning steps and function-calling outputs, which guide subsequent actions in the simulation loop. This enables modular integration across different models and circuit types.}
\label{fig：Communication}
\end{figure}

\subsection{LLM APIs as modular elements}

In EEsizer, the LLM APIs are used as a modular element within the agent system. The agent communicates with different LLMs by constructing structured prompts and passing them through the API, which then returns responses that guide subsequent actions in the simulation loop. The process is shown in Fig.~\ref{fig：Communication}. This design ensures that the same agent logic can be applied across different LLMs and different circuit types. It also enables systematic benchmarking of models such as Gemini~2.0 Flash, GPT-4.1, and OpenAI~o3, allowing fair comparison of their optimization performance in terms of number of iterations, success rate, and robustness.

\section{Results}

\subsection{Initial Test}
\begin{figure*}[t]
\centerline{\includegraphics[width=1\textwidth]{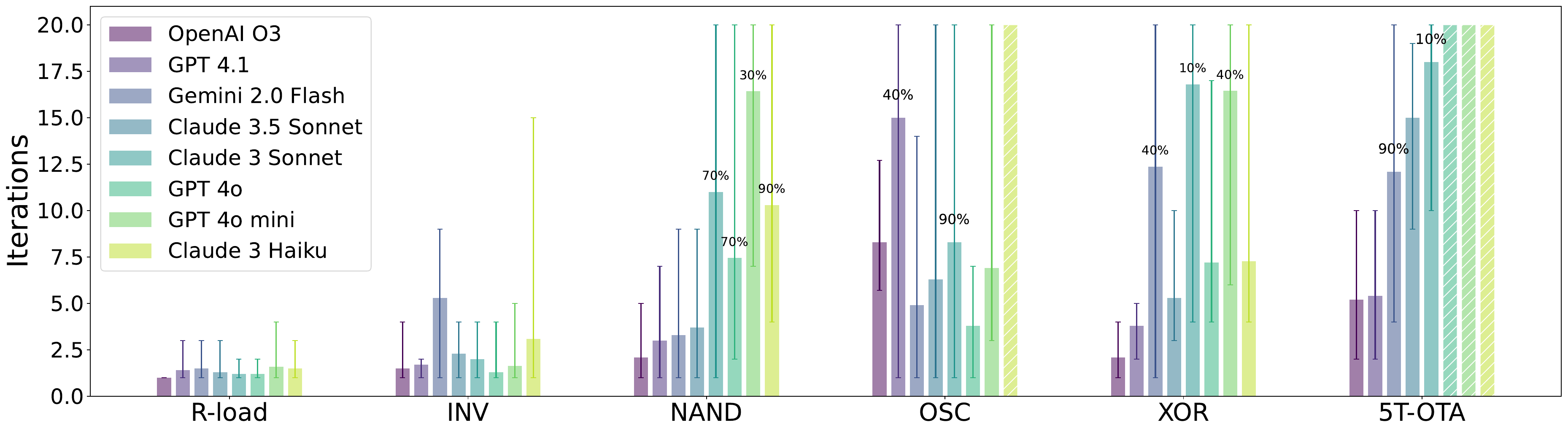}}
\caption{Performance comparison of different LLMs. This chart compares the performance of various LLMs across different circuits within 20 iterations. The height of the bars represents the average number of iterations, with lower values indicating better optimization performance. Error bars show the range of iterations observed across ten attempts. In each attempt, exceeding 20 iterations is considered a failure. The success rate smaller than 100\% is labelled on the bar. Hatched bars indicate that the model failed to optimize specific circuits in all ten attempts.}
\label{fig: performance}
\end{figure*}

In this work, we extend \cite{11107079} to evaluate eight LLMs: Anthropic’s Claude 3.5 Sonnet, Claude 3 Sonnet, Claude 3 Haiku; OpenAI’s GPT-4o mini, GPT-4o, GPT-4.1, o3; and Google DeepMind’s Gemini 2.0 Flash across six different circuits. Each model was assessed using identical prompts, functions, and performance metrics. The evaluation focused on the number of iterations required to achieve successful optimization, with a maximum limit of 20 iterations per run. All the tests used a PTM 180nm transistor model and initialized with W=1~µm, L=180~nm. All the tests were
performed in August of 2025.

The performance metrics targeted for each circuit are listed in Table.~\ref{tab:metrics} and the results are summarized in Fig.~\ref{fig: performance}. OpenAI~o3 exhibited the best overall performance, achieving a 100\% success rate across all circuits with the fewest iterations and the lowest variability, despite requiring relatively more iterations on the ring oscillator. Claude~3.5 Sonnet also achieved a 100\% success rate, however, it required noticeably more iterations on the 5T-OTA circuit. GPT-4.1 performed comparably on most circuits, however, it achieved only a 40\% success rate on the ring oscillator. Gemini 2.0 Flash performed worse on the XOR gate, however, it achieved higher success rates on the other circuits, despite exhibiting larger variations. Claude~3 Sonnet and GPT-4o mini exhibited similar trends, with lower success rates and higher variability, particularly for the NAND and XOR gates. Claude~3 Haiku succeeded on digital circuits, however, it consistently failed on the oscillator and 5T-OTA. 

\renewcommand{\arraystretch}{1.2}
\begin{table}[t]
\centering
\caption{Target performance specifications used for the initial LLMs evaluation across different circuits. All models were tested using these metrics.}
\resizebox{0.5\textwidth}{!}{%
    \begin{tabular}{c|c|c|c|c|c}
        \hline
        \textbf{Circuit} & \textbf{Type} & \makecell{\textbf{Num.}\\\textbf{Transistors}} & \multicolumn{3}{c}{\textbf{Target Performance}} \\
        \hline
        \hline
        \multirow{2}{*}{\textbf{R-load}} & \multirow{5}{*}{\textbf{Analog}} & \multirow{2}{*}{1} & \textbf{Gain (dB)} & \textbf{Bandwidth (MHz)} & \textbf{Phase Margin} \\
        \cline{4-6}
        & & & 20 & 1 & 60 \\
        \cline{1-1} \cline{3-6}
        \textbf{OTA} & & 7 & 40 & 1 & 60 \\
        \cline{1-1} \cline{3-6}
        \multirow{2}{*}{\textbf{OSC}} & & \multirow{2}{*}{6} & \multicolumn{3}{c}{\textbf{Frequency (GHz)}} \\
        \cline{4-6}
        & & & \multicolumn{3}{c}{10} \\
        \hline
        \hline
        \multirow{2}{*}{\textbf{INV}} & \multirow{4}{*}{\textbf{Digital}} & \multirow{2}{*}{2} & \textbf{Delay (ps)} & \multicolumn{2}{c}{\textbf{Transient Power (pW)}} \\
        \cline{4-6}
        & & & 10 & \multicolumn{2}{c}{5} \\
        \cline{1-1} \cline{3-6}
        \textbf{NAND} & & 4 & 20 & \multicolumn{2}{c}{5} \\
        \cline{1-1} \cline{3-6}
        \textbf{XOR} & & 12 & 30 & \multicolumn{2}{c}{5} \\
        \hline
    \end{tabular}%
}
\label{tab:metrics}
\end{table}

\begin{figure}[t]
    \centering
    \resizebox{0.5\textwidth}{!}{%
        \includegraphics{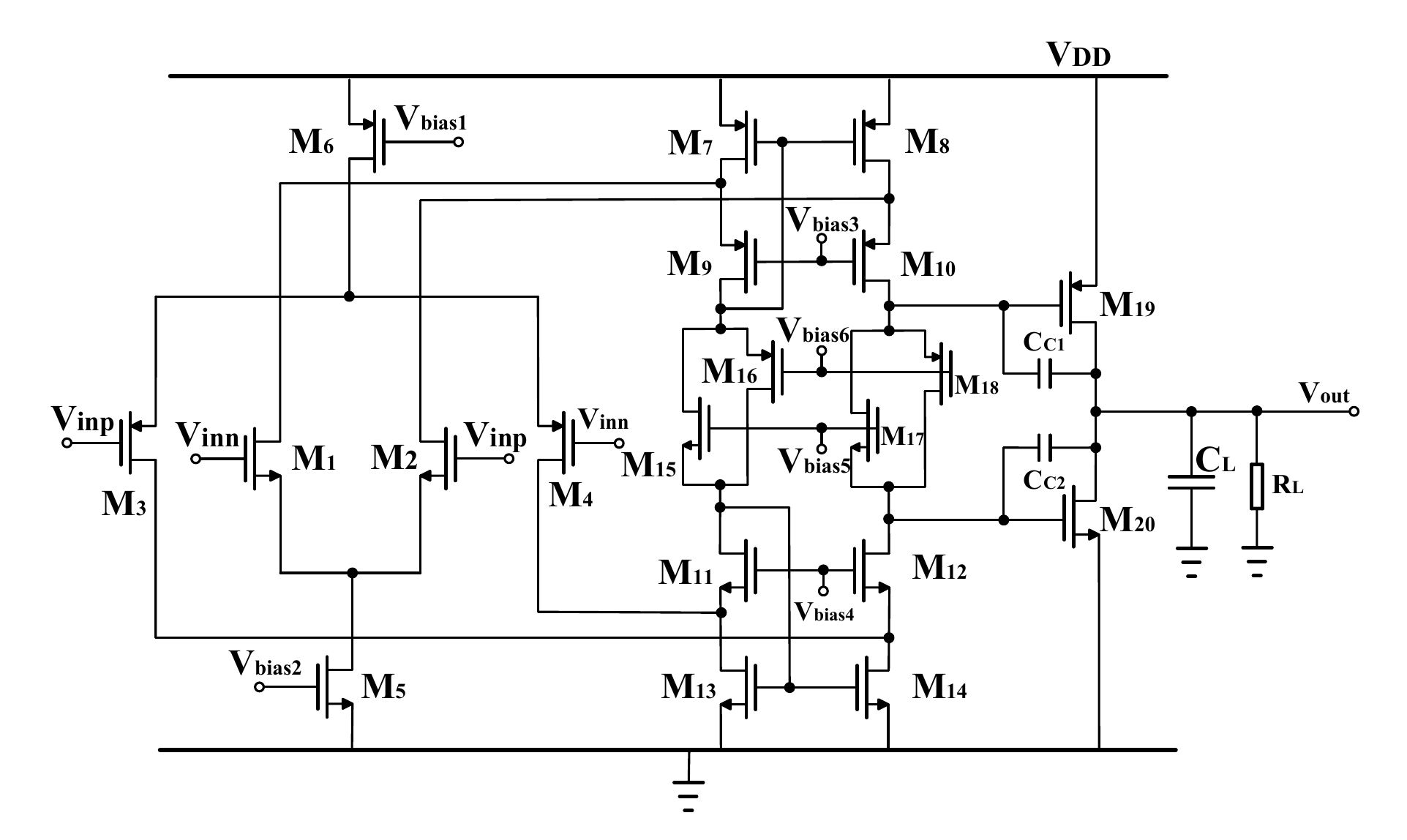}
    }
    \caption{Schematic of the tested opamp, featuring a complementary input stage and a class AB output stage.
}
    \label{fig:schematic}
\end{figure}

\subsection{The Performance of Different LLMs on the OpAmp}
Motivated by the above findings, we focus on the best-performing LLMs OpenAI~o3 and GPT-4.1, which excelled on the optimization of 5T-OTA, along with the top mid-sized model, Gemini~2.0 Flash, to optimize a complementary input stage and a class-AB output stage consisting of 20 transistors ~\cite{hogervost1994compact}. The schematic is shown in Fig.~\ref{fig:schematic}. In all tests, 10 performance metrics, including rail-to-rail operation and load conditions, are required. In the experiment, Gemini~2.0 Flash, GPT-4.1, and OpenAI~o3 were separately used to optimize the opamp on PTM 180~nm, 130~nm, and 90~nm technology nodes. All the tests were
performed in August of 2025.

\begin{figure}[t]
    \centering
    \resizebox{0.5\textwidth}{!}{%
        \includegraphics{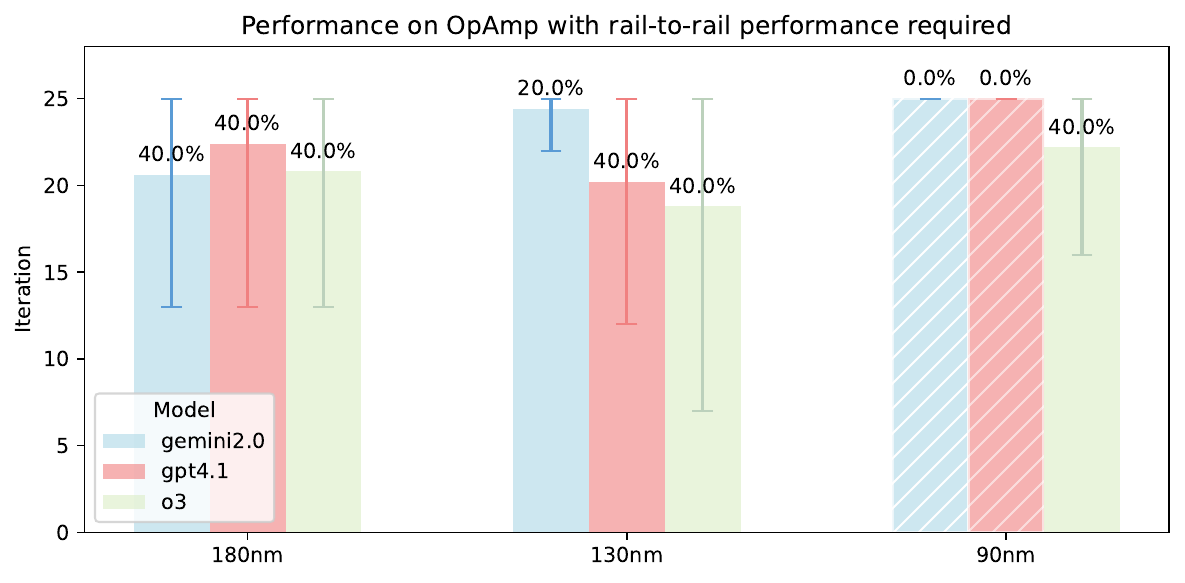}
    }
    \caption{Performance comparison of different LLMs on opamp optimization on PTM 180~nm, 130~nm and 90~nm. This chart compares the performance of 3 LLMs. Error bars show the range of iterations observed across five attempts. In each attempt, exceeding 25 iterations is considered a failure. The success rate smaller than 100\% is labelled on the bar. Hatched bars indicate that the model failed to optimize specific circuits in all attempts.}
    \label{fig:r2r_performance}
\end{figure}

\begin{table}[t]
\centering
\caption{Computation efficiency Comparison between Different LLMs.}
\renewcommand{\arraystretch}{1.2}
\resizebox{0.48\textwidth}{!}{%
\begin{tabular}{c|c|c|c|c}
\hline
\textbf{Model} & 
\makecell{\textbf{Average} \\ \textbf{Iterations}}&
\makecell{\textbf{Total} \\ \textbf{Optimization} \\ \textbf{Time}} & 
\makecell{\textbf{Optimization} \\ \textbf{Time per} \\ \textbf{Iteration}} & 
\makecell{\textbf{Simulation } \\ \textbf{Time per} \\ \textbf{Iteration}} \\
\hline
\hline
OpenAI o3 &22  & 73m 27.43s & 3m 15.83s & 3.98s \\

GPT 4.1  &25 & 60m 35.26s & 2m 27.81s & 4.01s \\

Gemini 2  &25 & 38m 44.08s & 1m 32.96s & 3.90s \\

\hline
\end{tabular}%
}
\label{tab:optimization_time}
\end{table}

Fig.~\ref{fig:r2r_performance} presents the performance of different LLMs across 180~nm, 130~nm, and 90~nm technology nodes. For each node, five independent attempts were conducted. As the technology scales down from 180~nm to 90~nm, the success rate decreases for all three LLMs. 

For the LLMs evaluated, o3 exhibited the best overall performance, achieving the highest success rate, the fewest iterations, and the lowest variability, indicating robust stability in circuit sizing. The GPT-4.1 performed comparably on the 180~nm and 130~nm nodes but failed on the 90~nm node. Gemini~2.0 Flash demonstrated superior performance on the 180~nm node but underperformed on the 130~nm and 90~nm nodes. Notably, for the 90~nm technology, only o3 successfully optimized the circuits, highlighting that models trained on larger datasets and with more parameters exhibit stronger reasoning capabilities, enabling more stable and efficient optimization.

As shown in Table~\ref{tab:optimization_time}, the computation time is dominated by the LLM reasoning and communication overhead during each optimization loop. The simulation time per iteration remains nearly constant across all models, indicating that SPICE-level simulation contributes minimally to the overall computation time. Among the evaluated models, Gemini 2.0 achieves the shortest total optimization time, demonstrating high computational efficiency but a lower success rate. In contrast, OpenAI o3 requires the longest total time yet delivers the most reliable optimization results. GPT-4.1 lies between these two, maintaining a balance between reasoning thoroughness and speed. It is important to note that these performance characteristics are dependent on the specific LLM provider and are subject to change with future model updates.

\subsection{The Performance on Different Technology Nodes}
Table~\ref{tab:performance_nodes} presents the results of opamp optimization across 180~nm, 130~nm, and 90~nm technology nodes by LLM Gemini~2.0 Flash. Each node was tested with five independent optimization attempts starting from the same initial conditions and target specifications. A load capacitor $C_\text{L} = 10\,\text{pF}$ and load resistor $R_\text{L} = 1\,\text{k}\Omega$ were used for all tests, and a 5\% tolerance was applied to all performance metrics. Iterations exceeding 25 without meeting the target were considered failures.

As the technology scales from 180~nm to 90~nm, the number of failed attempts increases, reflecting stronger short-channel effects and tighter matching constraints. In all the tests, input offset, output swing and input common-mode range improved significantly, however, they failed in many of the tests, showing the difficulty in achieving rail-to-rail requirements. Transient output performance, measured as total harmonic distortion (THD), exhibits a clear degradation trend with node scaling. In addition, with the same capacitive and resistive load, advanced nodes exhibit more failures in small-signal performance, including gain and phase margin, suggesting higher load sensitivity. Convergence also becomes less efficient, as most successful attempts at 180~nm and 130~nm complete within 22 iterations, whereas 90~nm rarely converges within the iteration budget.

\begin{table}[t!] 
\centering
\renewcommand{\arraystretch}{1.2}
\caption{Performance evaluation for five attempts across three technology nodes: 180~nm, 130~nm, 90~nm. For all the tests, a load capacitor $C_L = 10\,\text{pF}$ and load resistor $R_L = 1\,\text{k}\Omega$ is considered. A 5\% tolerance is applied to all metrics. If the 25th iteration is reached without meeting any of the targets, it is marked as a failure.}
\resizebox{0.5\textwidth}{!}{ 
\begin{tabular}{c|c|c|c|c|c|c|c|c|c|c}
\hline
\textbf{Node} 
& \textbf{Iter.}   
& \makecell{\textbf{Gain}\\\textbf{(dB)}}     
& \makecell{\textbf{UGBW}\\\textbf{(MHz)}} 
& \makecell{\textbf{PM}\\\textbf{($^\circ$)}} 
& \makecell{\textbf{Power}\\\textbf{(mW)}} 
& \makecell{\textbf{CMRR}\\\textbf{(dB)}}     
& \makecell{\textbf{THD}\\\textbf{(dB)}}   
& \makecell{\textbf{Offset}\\\textbf{(mV)}}   
& \makecell{\textbf{Output}\\\textbf{Swing}\\\textbf{(V)}} 
& \makecell{\textbf{ICMR}\\\textbf{(V)}} \\
\hline
\hline

\textcolor{blue}{\textbf{Target}} & \textcolor{blue}{25} & \textcolor{blue}{$\geq65$} & \textcolor{blue}{$\geq10$} & \textcolor{blue}{$\geq50$} & \textcolor{blue}{$\leq10$} & \textcolor{blue}{$\geq100$} & \textcolor{blue}{$\leq-26$} & \textcolor{blue}{$\leq1$} & \textcolor{blue}{Vdd-Vss} & \textcolor{blue}{Vdd-Vss} \\
\hline
\hline

\multirow[c]{6}{*}{180nm} 
& \textbf{ Initial} & \textbf{31.17} & \textbf{0.32} & \textbf{85.52} & \textbf{0.53} & \textbf{95.60} &     \textbf{-27.06} & \textbf{587} & \textbf{0.13} & \textbf{0.03} \\
\cline{2-11}
 & \textbf{13} & 63.99 & 12.59 & 53.55 & 9.73 & 127.09 & -26.04   & 0.16 & 1.72  & 1.75 \\
 & \textbf{15} & 62.53 & 50.12 & 49.53 & 5.10 & 120.67  & -27.29   & 0.46 & 1.71  & 1.72 \\
 & fail & \textcolor{red}{28.05} & \textcolor{red}{0.4} & 86.02 & 4.12 & 100.09 & -25.95   & 0.09 & 1.76  & 1.79 \\
 & fail & 65.05 & 10.00 & \textcolor{red}{42.19} & 1.7  & 141.53 & -25.13   & \textcolor{red}{6.32} & \textcolor{red}{1.42}  & \textcolor{red}{1.42} \\
 & fail & 62.41 & 15.01 & 50.27 & 0.94 & 113.23 & -26.21   & 0.22 & \textcolor{red}{1.60}  & \textcolor{red}{1.63} \\
\hline
\hline

\multirow[c]{6}{*}{130nm} 
& \textbf{ Initial} & \textbf{33.76} & \textbf{10.00} & \textbf{48.09} & \textbf{1.72} & \textbf{80.69} &     \textbf{-49.59} & \textbf{235} & \textbf{0.38} & \textbf{0.43} \\
\cline{2-11}
 
 & \textbf{22} & 62.52 & 10.00 & 47.68 & 3.90 & 139.00  & -24.76  & 0.03 & 1.72  & 1.75 \\
 & fail & 74.06 & 39.81 & 68.86 & \textcolor{red}{14.66}  & 132.13 & \textcolor{red}{-24.61} & 0.07 & 1.76  & 1.79 \\
 & fail & 65.67 &39.81 & \textcolor{red}{36.19} & 2.96 & 105.67 & -29.76  & 0.13 & \textcolor{red}{1.34}  & \textcolor{red}{1.37} \\
 & fail & 73.80 & 100.00 & 63.89 & 10.46 & 105.84 & \textcolor{red}{-18.25}  & 0.008 & \textcolor{red}{1.60}  & \textcolor{red}{1.63} \\
 & fail & \textcolor{red}{34.06} & 39.81 & \textcolor{red}{35.60} & \textcolor{red}{12.33} & \textcolor{red}{50.72} & -29.31   & \textcolor{red}{5.43} & \textcolor{red}{1.63}  & \textcolor{red}{1.21} \\
\hline
\hline
\multirow[c]{5}{*}{90nm}
& \textbf{ Initial} & \textbf{33.23} & \textbf{0.50} & \textbf{86.55} & \textbf{0.68} & \textbf{55.93} &     \textbf{-28.35} & \textbf{176} & \textbf{0.21} & \textbf{0.31} \\
\cline{2-11}
 & fail & \textcolor{red}{38.64} & 79.43 & \textcolor{red}{30.61} & 7.91 & \textcolor{red}{79.09} & -40.28 & \textcolor{red}{1.96} & \textcolor{red}{1.06} & \textcolor{red}{1.10} \\
 & fail & 82.07 & 125.89 & \textcolor{red}{31.38} & 2.70  & 137.86 &  \textcolor{red}{-2.94}   & 0.01 & 1.17 & 1.19 \\
 & fail & 68.10 & 63.10 & 55.29 & 2.99 & 134.84 & \textcolor{red}{-10.47}   & 0.03 & \textcolor{red}{1.15}  & \textcolor{red}{1.18} \\
 & fail & 78.57 & 251.19 & 76.79 & \textcolor{red}{10.8} & 126.52 & \textcolor{red}{-9.55}   & 0.02  & 1.15 & 1.17 \\
 & fail & \textcolor{red}{55.02} & 31.62 & 63.49 & 4.42 & 102.12 & \textcolor{red}{-17.55}    & 0.26 & 1.17 & 1.19 \\
\hline
\end{tabular}}
\label{tab:performance_nodes}
\end{table}

\subsection{The Performance on Different Groups of Metrics}
We set 10 performance metrics of operational amplifiers across three configuration groups. In all groups, rail-to-rail operation is required, with targets set for the full input common-mode range and output swing. The groups, however, differ in bandwidth, load conditions, and power consumption requirements to emulate realistic application scenarios. As an illustrative example, tests were conducted using OpenAI~o3 on the PTM 90~nm technology. Table~\ref{tab:performance_metrics} summarizes the performance metrics for each group, including the final results, number of passed and failed attempts.

\begin{table}[t!] 
\centering
\renewcommand{\arraystretch}{1.1}
\caption{Performance evaluation for five attempts across three groups (G1, G2 and G3). A 5\% tolerance is applied to all metrics; deviations beyond this tolerance are marked in \textcolor{red}{red}. Different targets for G1 and G2, G1 and G3 are marked in \textbf{bold}. For G1 and G3, a load capacitor $C_L = 10\,\text{pF}$ and load resistor $R_L = 1\,\text{k}\Omega$ is considered. For G2, a load capacitor $C_L = 50\,\text{pF}$ and load resistor $R_L = 100\,\text{k}\Omega$ is consideredIf the 25th iteration is reached without meeting any of the targets, it is marked as a failure.}
\resizebox{0.5\textwidth}{!}{ 
\begin{tabular}{c|c|c|c|c|c|c|c|c|c|c}
\hline
\textbf{Group} 
& \textbf{Iter.}   
& \makecell{\textbf{Gain}\\\textbf{(dB)}}     
& \makecell{\textbf{UGBW}\\\textbf{(MHz)}} 
& \makecell{\textbf{PM}\\\textbf{($^\circ$)}} 
& \makecell{\textbf{Power}\\\textbf{(mW)}} 
& \makecell{\textbf{CMRR}\\\textbf{(dB)}}     
& \makecell{\textbf{THD}\\\textbf{(dB)}}   
& \makecell{\textbf{Offset}\\\textbf{(mV)}}   
& \makecell{\textbf{Output}\\\textbf{Swing}\\\textbf{(V)}} 
& \makecell{\textbf{ICMR}\\\textbf{(V)}}\\
\hline
\hline
\textcolor{blue}{G1 Target} & \textcolor{blue}{25} & \textcolor{blue}{$\geq65$} & \textcolor{blue}{$\geq10$} & \textcolor{blue}{$\geq50$} & \textcolor{blue}{$\leq10$} & \textcolor{blue}{$\geq100$} & \textcolor{blue}{$\leq-26$} & \textcolor{blue}{$\leq1$} & \textcolor{blue}{1.2} & \textcolor{blue}{1.2} \\
\hline
\textbf{G1 Initial}& \textbf{0} & \textbf{33.23} & \textbf{0.50} & \textbf{86.55} & \textbf{0.68} & \textbf{55.93} &     \textbf{-28.35} & \textbf{176} & \textbf{0.21} & \textbf{0.31} \\
\hline
G1-1 & \textbf{16} & 69.74 & 25.12 & 70.74 & 6.08 & 102.36 & -36.35 & 0.03 & 1.15 & 1.19  \\
G1-2 & \textbf{20} & 69.14 & 19.95 & 77.01 & 6.11 & 112.00 & -36.82 & 0.02 & 1.16 & 1.19  \\
G1-3 & fail & \textcolor{red}{59.21} & 79.43 & 76.57 & 0.61 & \textcolor{red}{23.17} & -27.95 & 0.52 & \textcolor{red}{0.95} & \textcolor{red}{1.02}  \\
G1-4 & fail & \textcolor{red}{53.60} & 39.81 & 79.82 & 4.66 & \textcolor{red}{36.28} & -25.52 & 0.18 & 1.15 & 1.19  \\
G1-5 & fail & \textcolor{red}{34.09} & 10.00 & 61.97 & 1.74 & \textcolor{red}{70.17} & -42.05 & \textcolor{red}{11.34} & \textcolor{red}{0.85} & \textcolor{red}{0.97} \\
\hline
\hline

\textcolor{blue}{G2 Target} & \textcolor{blue}{25} & \textcolor{blue}{$\geq65$} & \boldmath{\textcolor{blue}{$\geq 5$}} & \boldmath{\textcolor{blue}{$\geq 45$}} & \boldmath{\textcolor{blue}{$\leq 5$}} & \textcolor{blue}{$\geq100$} & \textcolor{blue}{$\leq-26$} & \textcolor{blue}{$\leq1$} & \textcolor{blue}{1.2} & \textcolor{blue}{1.2} \\
\hline
\textbf{G2 Initial}& \textbf{0} & \textbf{40.03} & \textbf{1} & \textbf{75.05} & \textbf{0.56} & \textbf{46.61} &     \textbf{-37.91} & \textbf{2.05} & \textbf{1.08} & \textbf{1.19} \\
\hline
G2-1 & \textbf{20} & 65.83 & 12.59 & 61.26 & 1.50 & 115.15 & -25.77 & 0.01 & 1.19 & 1.19 \\
G2-2 & fail & 66.36 & \textcolor{red}{1.99} & 48.06 & 0.40 & 107.62 & -31.71 & 0.06 & \textcolor{red}{1.04} & 1.15    \\
G2-3 & fail & \textcolor{red}{47.04} & \textcolor{red}{2.51} & \textcolor{red}{36.19} & 0.40 & \textcolor{red}{84.85} & -26.03 & 0.48 & 1.18 & 1.19  \\
G2-4 & fail & \textcolor{red}{53.87} & 5.01 & \textcolor{red}{42.23} & 1.52 & 98.18 & -25.62 & 0.12 & 1.19 & 1.19  \\
G2-5 & fail & \textcolor{red}{57.23} & 6.31 & \textcolor{red}{25.68} & 0.50 & \textcolor{red}{46.76} & \textcolor{red}{-23.41} & 0.14 & 1.19 & 1.19  \\
\hline
\hline

\textcolor{blue}{G3 Target} & \textcolor{blue}{25} & \textcolor{blue}{$\geq65$} & \boldmath{\textcolor{blue}{$\geq50$}} & \boldmath{\textcolor{blue}{$\geq50$}} & \boldmath{\textcolor{blue}{$\leq20$}} & \boldmath{\textcolor{blue}{$\geq80$}} & \boldmath{\textcolor{blue}{$\leq-26$}} & \textcolor{blue}{$\leq1$} & \textcolor{blue}{1.2} & \textcolor{blue}{1.2}  \\
\hline
\textbf{G3 Initial} & \textbf{0} & \textbf{33.23} & \textbf{0.50} & \textbf{86.55} & \textbf{0.68} & \textbf{55.93} & \textbf{-28.35} & \textbf{176} & \textbf{0.21} & \textbf{0.31}  \\
\hline
G3-1 & \textbf{16} & 64.51 & 125.89 & 68.75 & 6.58 & 103.68 & -32.60 & 0.24 & 1.14 & 1.17\\

G3-2 & fail & 66.67 & \textcolor{red}{39.81} & \textcolor{red}{37.09} & 4.23 & 103.50 & \textcolor{red}{-21.05} & 0.69 & \textcolor{red}{1.13} & 1.18  \\

G3-3 & fail & 75.34 & 125.89 & \textcolor{red}{33.74} & 13.72 & 126.24 & \textcolor{red}{-20.15} & 0.06 & 1.17 & 1.19  \\
G3-4 & fail & 67.68 & 63.09 & 51.55 & 3.69 & 120.84 & \textcolor{red}{-18.39} & 0.09 & \textcolor{red}{1.11} & 1.14  \\
G3-5 & fail  & 66.67  & \textcolor{red}{37.08}  & \textcolor{red}{39.81}  & 4   & 103.51  & \textcolor{red}{-21.08} & 0.07 &1.14  &1.18  \\
\hline
\end{tabular}}
\label{tab:performance_metrics}
\end{table}

In G1, where the opamp is loaded by a heavy resistive component, the agent consistently achieved the unity-gain bandwidth and phase margin specifications across all attempts, thanks to the relatively lower capacitive load of 10~pF. The remaining three attempts failed primarily due to the rail-to-rail input and output requirement.

In G2, where a heavier capacitive load of 50~pF and stricter power budget were imposed, optimization success was notably more difficult, especially on small-signal frequency response. Rail-to-rail performance in this case is not a strong limit because of the lower resistive load, which makes the input and output swing initially near the full supply rail.

In G3, where the bandwidth target was most aggressive, gain was easily achieved, however, offset, bandwidth, and phase margin degraded, reflecting the inherent trade-off between high bandwidth operation and precision. Additionally, THD failed the most, indicating that aggressively pushing the gain-bandwidth can exacerbate distortion and stability issues.

\begin{figure}[t]
    \centering
    \resizebox{0.5\textwidth}{!}{%
        \includegraphics{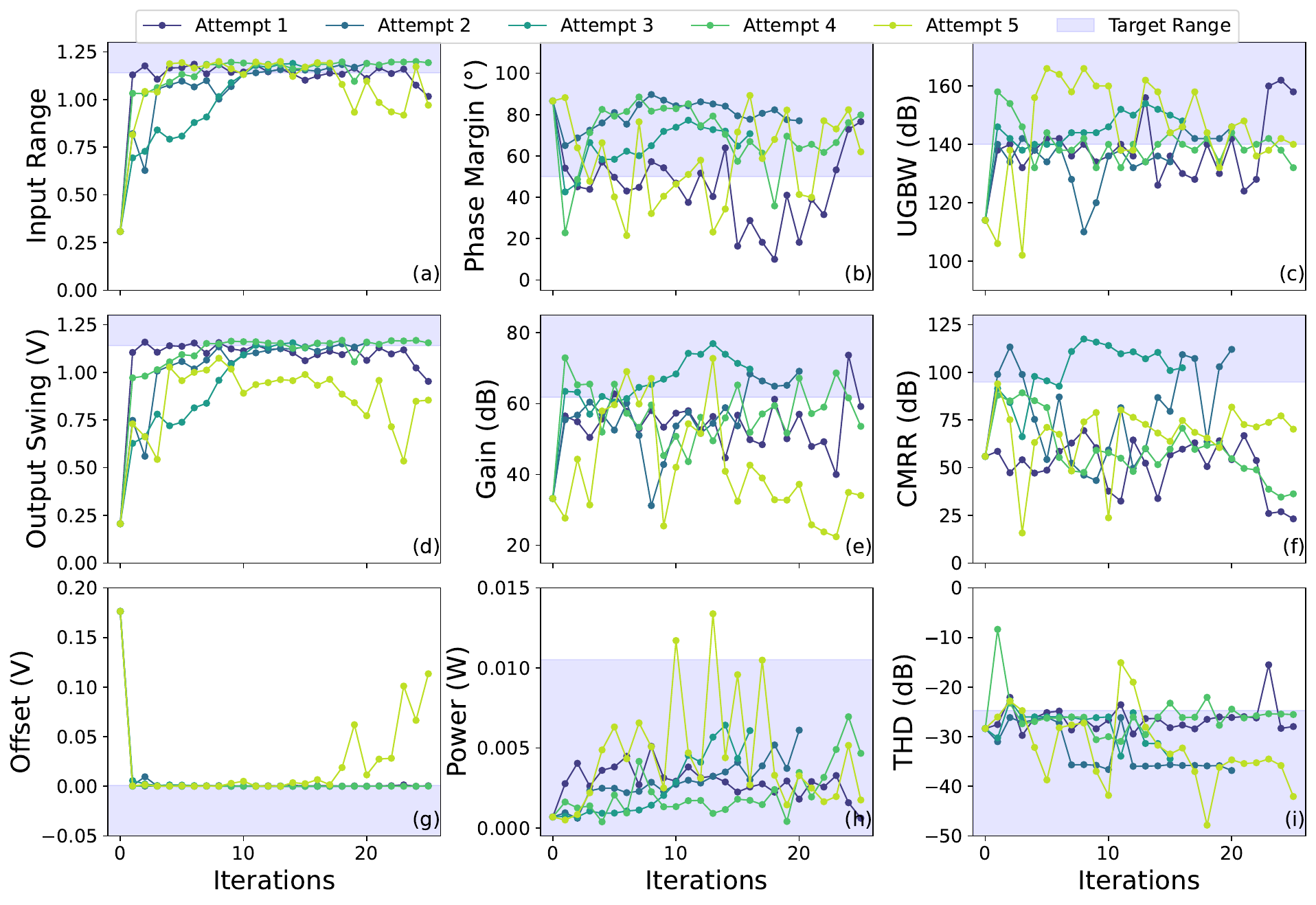}
    }
    \caption{Optimization results for the opamp. (a) Input Common-Mode Range (b) Phase Margin (c) Unity-Gain Bandwidth (d) Output Swing (e) Gain (f) CMRR (g) Input Offset (h) Power (i) Total Harmonic Distortion.}
    \label{fig:4*2}
\end{figure}

The optimization process of the opamp under G1 is illustrated in Fig.~\ref{fig:4*2}. Although performance metrics fluctuate, the agent successfully refined its strategy to meet all specifications in two out of five attempts. The variation across trials arises from the dynamic interaction between the LLM’s probabilistic reasoning and the circuit simulation feedback. In each optimization round, the outcome of the performance calculations guides the LLM’s subsequent parameter adjustments. Failed trials typically occur when the agent explores suboptimal regions of the design space or makes fewer effective decisions during exploration.

\begin{table}[h!]
\centering
\caption{Transistor size and bias voltages for G1-2, G2-5 and G3-5 after optimization}
\resizebox{0.5\textwidth}{!}{
\begin{tabular}{c|c|c|c}
\hline
\textbf{Transistor} & \textbf{G1-2 (W/L, \text{µm}/\text{µm})} & \textbf{G2-1 (W/L, \text{µm}/\text{µm})} &  \textbf{G3-1 (W/L, \text{µm}/\text{µm})} \\ \hline
M1, M2   & 32/0.85    & 20/0.70 & 12/0.25 \\ 
M3, M4   & 72/0.85    & 40/0.70 & 24/0.25\\ 
M5       & 3.5/0.50   & 4/0.24 & 12/0.20\\ 
M6       & 10/0.80    & 22/0.30 & 24/0.25 \\ 
M7, M8   & 6/0.18     & 24/0.45 & 30/0.35\\ 
M9, M10  & 6/0.18     & 24/0.45 & 30/0.35\\ 
M11, M12 & 3.5/1.20   & 6/0.24 & 6/0.25\\ 
M13, M14 & 2/1.20     & 1/0.09 &  6/0.30 \\ 
M15, M17 & 3/0.22     & 1/0.12 & 6/0.22\\
M16, M18 & 10/1.4     & 5/0.15 & 24/0.30\\ 
M19      & 280/0.38    & 100/0.70 & 130/0.12 \\ 
M20      & 140/0.38    & 50/0.70 & 65/0.90 \\ 
\hline
\hline
bias1 & 0.65  & 0.56 & 0.75 \\
bias2 & 0.80  & 0.60 & 0.67 \\
bias3 & 0.65  & 0.30 & 0.05 \\
bias4 & 0.80  & 1.03 & 0.90 \\
bias5 & 1.07  & 1.08 & 1.06 \\
bias6 & 0.075 & 0.14 & 0.07 \\
\hline
\end{tabular}
}

\label{tab:transistor_sizing}
\end{table}

The transistor sizes and bias voltages belonging to a successful attempt of each configuration group are shown in Table~\ref{tab:transistor_sizing}.

\begin{figure}[t]
    \centering
    \resizebox{0.5\textwidth}{!}{%
        \includegraphics{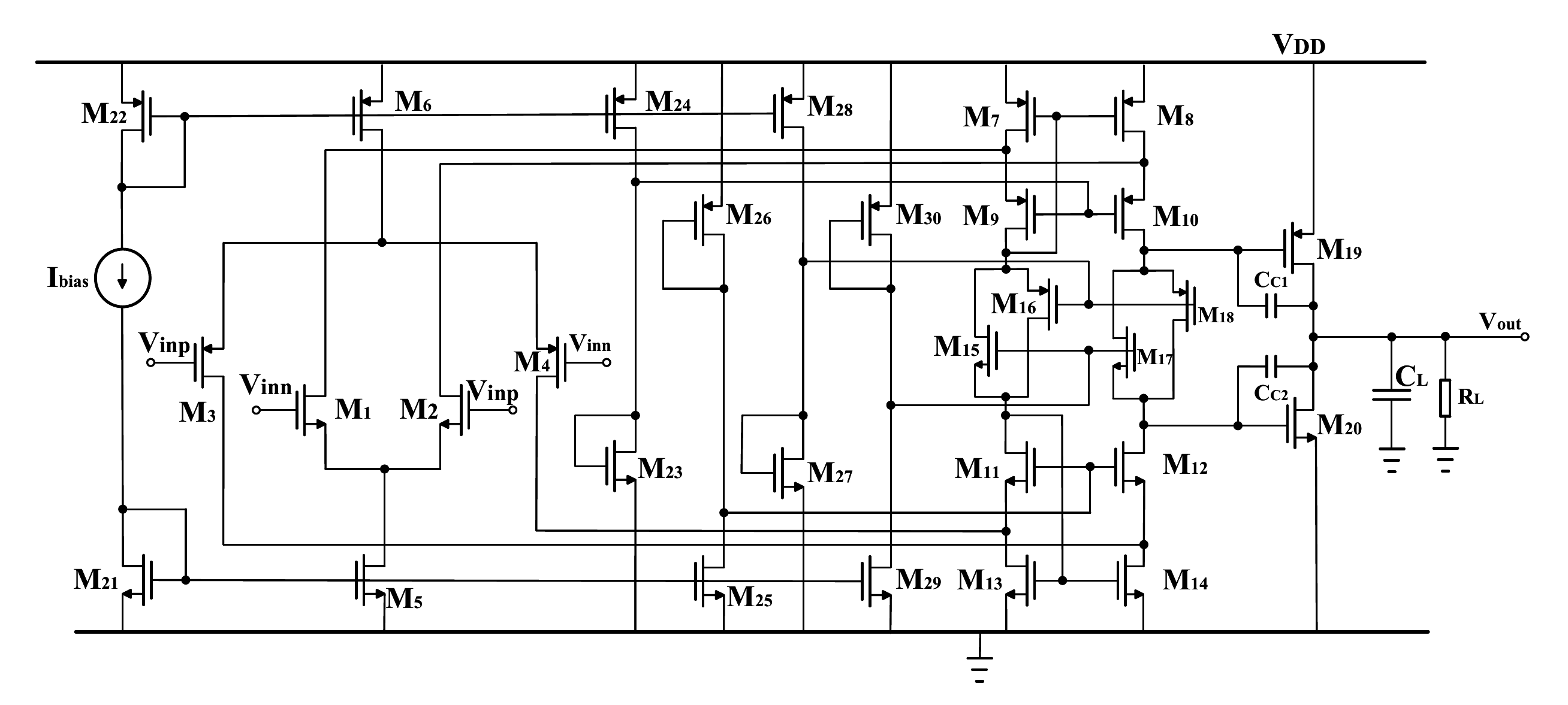}
    }
\caption{The complete circuit of Fig.~\ref{fig:schematic}, which includes a manually designed bias circuit targeting at providing bias voltages in Table.~\ref{tab:transistor_sizing}.}
\label{fig:current_source}
\end{figure}

\begin{figure}[t]
    \centering
    \resizebox{0.5\textwidth}{!}{%
        \includegraphics{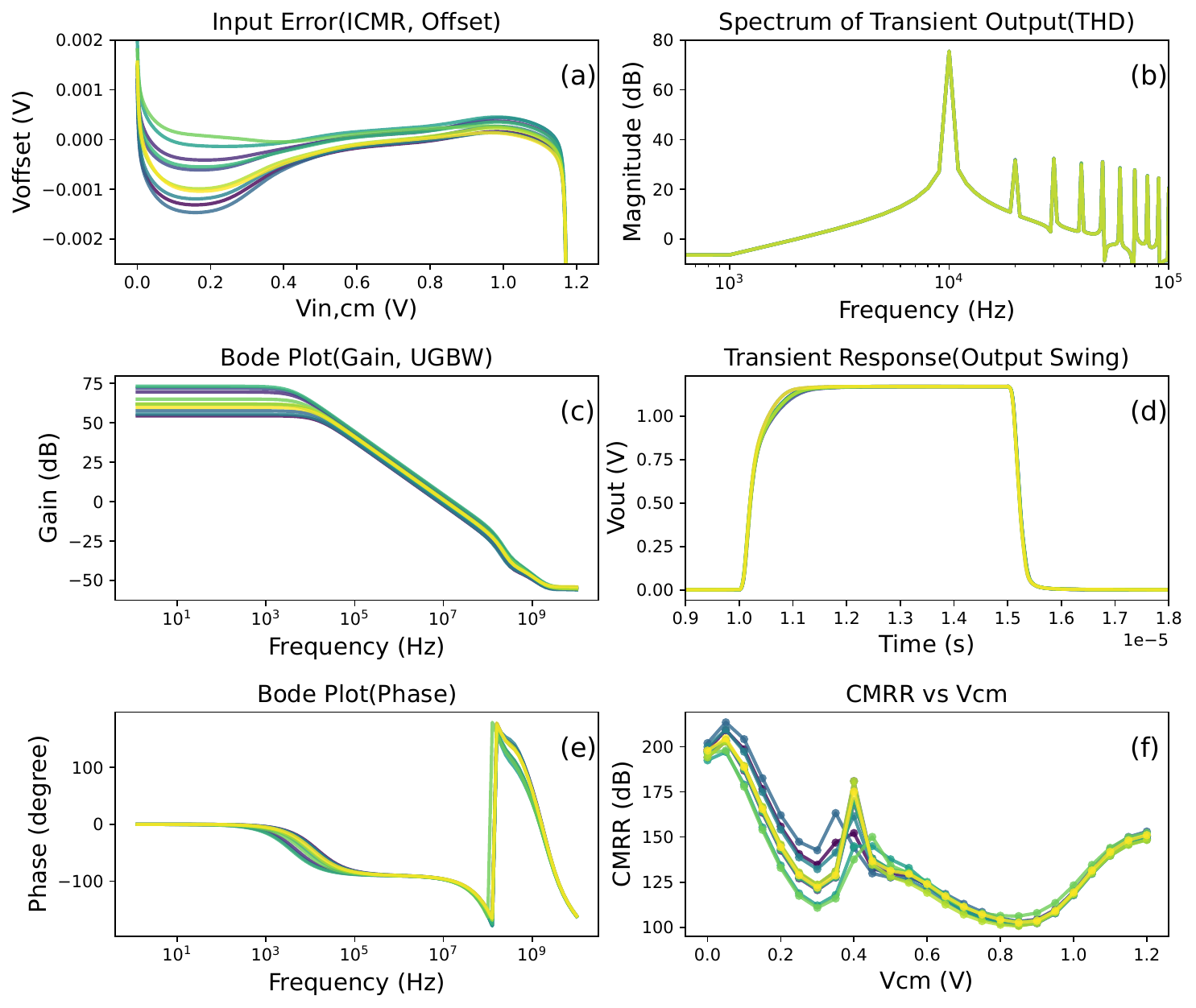}
    }
\caption{Variation results for G1-2 based on 10 samples. A standard Gaussian distribution with $\mu=0$, $\sigma=5$\,nm and $\mu=0$, $\sigma=10$\,mV, was performed on transistor size and $V_\text{th}$ respectively. (a) Input error across Vin,cm  (b) Spectrum of Transient Output (c), (e) Small-signal Bode plot (d) Transient output (f) CMRR across Vcm.}
\label{fig:opamp}
\end{figure}
\subsection{Variation Test}
To further assess the robustness of the optimized design, a bias circuit was manually implemented to provide the bias voltages listed in Table~\ref{tab:transistor_sizing}. A variation test was then performed on circuit G1-2 from Table~\ref{tab:performance_metrics}. In this test, device-level variations were introduced to both transistor dimensions and threshold voltage ({$V_\text{th}$). Transistor sizes were varied according to a Gaussian distribution with $\mu=0$, $\sigma=5$\,nm, while $V_\text{th}$ variations were modelled by adding a voltage source at the gate with $\mu=0$, $\sigma=10$\,mV. The complete circuit configuration, including bias circuits, is shown in Fig.~\ref{fig:current_source}.

The small-signal and large-signal simulation results under ten variation samples are summarized in Fig.~\ref{fig:opamp}. Despite device-level perturbations, DC and transient behaviors such as input offset and output swing remain stable, demonstrating rail-to-rail performance. The largest sensitivity is observed in gain, primarily due to variations in transconductance ($g_\text{m}$) and parasitic capacitances, which in turn affect the unity-gain bandwidth (UGBW) and overall frequency response.

To further quantify robustness, analysis was performed using 50 variation samples across all nine performance metrics, as shown in Fig.~\ref{fig:mc}. The results indicate that rail-to-rail metrics (ICMR and output swing) exhibit the smallest deviations and consistently meet specifications. In contrast, gain-related parameters (gain and UGBW) show larger spreads, with pass rates of 78\% and 76\%, respectively. 

\begin{figure}[t]
    \centering
    \resizebox{0.5\textwidth}{!}{%
        \includegraphics{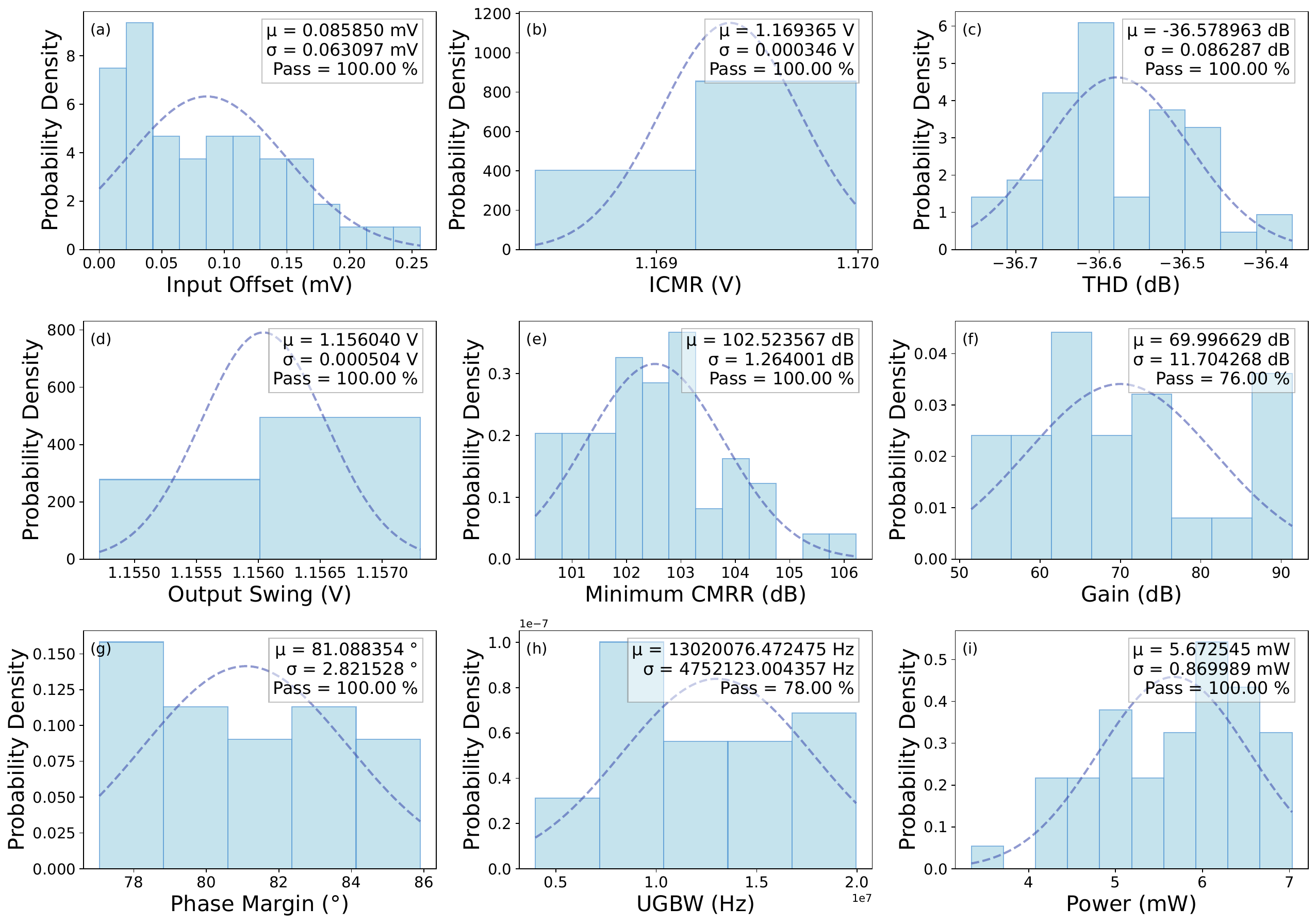}
    }
\caption{Variation results for G1-2 based on 50 samples. Variations were applied to transistor size and threshold voltage ($V_\text{th}$), following standard Gaussian distributions with $\mu=0$, $\sigma=5$\,nm and $\mu=0$, $\sigma=10$\,mV, respectively. The histograms represent the probability density, while the dashed lines indicate the Gaussian fit for each metric based on the calculated $\mu$ and $\sigma$. The performance that hits the target is considered a pass. The target performance group G1 is listed in Table.~\ref{tab:performance_metrics}.}
\label{fig:mc}
\end{figure}

\section{Discussion}
This work demonstrates that EEsizer can effectively generate circuit optimization strategies through a closed-loop optimization approach, leveraging both past and current performance feedback to balance multiple metrics. This highlights the potential of LLMs in the circuit sizing process. Nevertheless, several limitations remain.

First, layout and post-layout simulation were not considered in this study, as the focus was on transistor sizing. Incorporating area and layout constraints in future work would enable a more complete PPA trade-off.

Second, although a simplified variation test was performed on transistor size and $V_\text{th}$, a full Monte Carlo simulation with accurate device models incorporating Monte Carlo parameters should also be used to rigorously evaluate design robustness. As shown in Fig.~\ref{fig:mc}, not all performance metrics achieved a 100\% pass rate, highlighting the need to integrate Monte Carlo simulation into the agent’s workflow. This integration, will also eliminate the strong inversion requirement used in optimization loop to decide the current results pass or not in Section II-F. Furthermore, future studies should include noise performance. The PTM model we used may not include noise parameters, making accurate noise simulation infeasible. An accurate noise model will enable for an input referred noise and SNR performance measurement.

Third, the analysis functions for processing simulation results are currently designed manually. Since simulation settings and output conditions can vary significantly across different circuits, this limits generality and requires additional adaptation effort. Future work could explore leveraging LLMs for automatic analog testbench generation to improve adaptability.

Finally, optimization in our experiments was constrained to 25 iterations and five trials per configuration because of API rate and cost limits. These practical limits restrict the empirical coverage and make it difficult to fully characterize model behavior. Importantly, simply increasing the iteration budget does not guarantee better results: the LLM can oscillate between candidate solutions or repeatedly explore similar regions of the design space. Large, reasoning-focused LLMs such as OpenAI~o3 tend to produce higher-quality strategies and more stable proposals, however, their inference cost and latency make them unsuitable for dense, closed-loop experimentation over many trials. Furthermore, the success rate for the 20-transistor opamp appears low, due to the inherent complexity of the design constraints, the circuit topology, and LLM's limited expertise in analog circuit design. Although the success rate appears low, in the end, only one successful iteration is needed to achieve the goal of the sizing optimization.

\section{Conclusion}
We have developed EEsizer, an LLM-based AI agent that automates the transistor sizing process. The proposed closed-loop methodology enables data collection and annotation throughout optimization process, thereby minimizing manual effort for data collection. Combined with CoT prompt strategies that guide the LLMs step by step, the agent autonomously learns the trade-offs between transistor dimensions and performance, effectively serving as a sizing optimizer without domain-specific fine-tuning or reliance on external algorithms such as BO and gm/Id scripts. These capabilities, which were developed in this work, enable the agent to be adapted to various LLMs, circuits, specifications, and technology nodes without requiring modifications to prompts and functions. 
As a result, the proposed agent optimized six basic circuits, which include two kinds of amplifiers, a ring oscillator, and three logic gates, and further explored a 20-transistor opamp. Compared to \cite{11107079}, we demonstrated successful node transfer from 180~nm to 90~nm, covering ten performance metrics with rail-to-rail requirements using three LLMs. At 90~nm, OpenAI~o3 achieved the user-intended design targets across three test groups within 20 iterations, highlighting its adaptability at advanced nodes for analog circuits. Furthermore, a complete opamp design with a bias circuit was implemented, and 50 variations were performed on transistor size and $V_\text{th}$, achieving a minimum pass rate of 76\% for gain, which demonstrates the robustness of the proposed agent. In conclusion, this agent reflects the potential for optimizing more complex circuits and systems in the future. 

The source code for this project is available on https://github.com/eelab-dev.

\section*{Acknowledgements}

The authors thank EDINA and ISG@University of Edinburgh for their support in accessing OpenAI services. Also, Google Inc. for providing access to Gemini's AI services. We also thank Strategic Blue for enabling and advising on cloud-based AI services.

\bibliographystyle{IEEEtran}
\bibliography{IEEEabrv,bib}

\vspace{-0.5cm}
\begin{IEEEbiography}[{\includegraphics[width=1in,height=1.25in,clip,keepaspectratio]{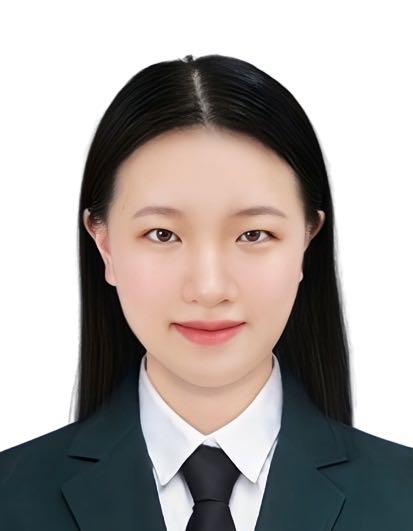}}]
{Chang Liu} received the B.E. degree from the School of Microelectronics at Hefei University of Technology, China, in 2023 and received her MSc in Electronics from The University of Edinburgh in 2024. She is now the recipient of Peter Denyer's PhD Scholarship at The University of Edinburgh. Her research focuses on leveraging LLMs to advance automation in AMS circuit design.

\end{IEEEbiography}

\vspace{-0.5cm}
\begin{IEEEbiography}[{\includegraphics[width=1in,height=1.25in,clip,keepaspectratio]{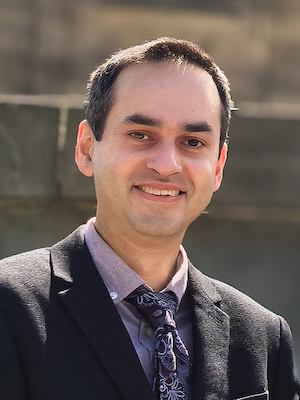}}]
{Danial Chitnis} received an MSc in Microelectronics at Bristol, UK, in 2008, and a DPhil in Engineering Science from the University of Oxford in 2013, and then joined the Department of Biomedical Physics at UCL, London, as a postdoctoral researcher. In 2017, he joined The University of Edinburgh as a Chancellor's Fellow in Electronics. His research focuses on using machine learning and artificial intelligence to improve the design and testing of integrated circuits and electronic systems.

\end{IEEEbiography}

\vfill

\end{document}